\documentclass[]{article}

\usepackage[preprint]{neurips_2024}

\usepackage[utf8]{inputenc} 
\usepackage[T1]{fontenc}    
\usepackage{hyperref}       
\usepackage{url}            
\usepackage{booktabs}       
\usepackage{amsfonts}       
\usepackage{nicefrac}       
\usepackage{microtype}      
\usepackage{xcolor}         
\usepackage{amsmath}
\usepackage{graphicx}
\usepackage{multirow}
\usepackage{algorithm}
\usepackage{algorithmic}
\usepackage{booktabs} 
\usepackage{array}    
\usepackage{tabularx}    
\usepackage{float}

\usepackage{siunitx}
\usepackage{adjustbox}

\title{Revisiting Data Compression with Language Modeling}

\author{%
  Chen-Han Tsai \\
  \texttt{maxwelltsai@yahoo.com} \\
}

\begin{document}

\maketitle

\begin{abstract}
In this report, we investigate the potential use of large language models (LLM's) in the task of data compression. Previous works have demonstrated promising results in applying LLM's towards compressing not only text, but also a wide range of multi-modal data. Despite the favorable performance achieved, there still remains several practical questions that pose a challenge towards replacing existing data compression algorithms with LLM's. In this work, we explore different methods to achieve a lower \textit{adjusted compression rate} using LLM's as data compressors. In comparison to previous works, we were able to achieve a new state-of-the-art (SOTA) adjusted compression rate of around $18\%$ on the enwik9 dataset without additional model training. Furthermore, we explore the use of LLM's in compressing non-English data, code data, byte stream sequences. We show that while LLM's excel in compressing data in text-dominant domains, their ability in compressing non-natural text sequences still remain competitive if configured in the right way. 

\end{abstract}

\section{Introduction}
Language modeling refers to the probabilistic process of approximating text distributions within natural language. The primary objective of a language model is to approximate the probability of a given text from its surrounding context. Although traditional rule-based language models (e.g., $n$-grams, hidden Markov models) have shown to be effective, their versatility and performance have not been able to match those achieved by recent deep learning-based Large Language Models (LLM's). LLM's are transformer-based language models \cite{transformers} that often contains billions of parameters. During training, LLM's are optimized to maximize the log-likelihood of each \textit{token}\footnote{For simplicity, we can consider \textit{tokens} as individual English words in this context. Additional discussions regarding tokenization will be introduced later.} based on a given text corpus. During inference, the LLM predicts the probability distribution of the next token for a given input token sequence. Studies \cite{scaling_laws} have shown that a large parameter size of an LLM is often indicative of better generalizability in predicting the next token correctly.

The predictive performance of an LLM makes it an ideal candidate for data compression. Several works \cite{deletang2024language, valmeekam2023llmzip, el_daher_connor_2006, li2024understandingcompression} have explored various methods to implement data compression using LLM's as a predictive model. The LLM first generates the probability distribution of each token, and these probability distributions are passed to an arithmetic encoder for encoding. To decode the sequence, the encoded data is passed sequentially into the LLM to predict the probability distribution of the next token. The symbol (or token) to decode can be determined by the interval of the predicted distribution for which the encoded data falls in. 

Previous works have demonstrated the strong compression abilities of LLM's in combination with arithmetic coding for text, image, and audio data. LLM-based compression methods have been compared to general purpose compressors \cite{gzip, pavlov20197z, welch1984technique} as well as domain-specific encoding standards \cite{boutell1997png, coalson2008flac}. Experimental results have shown that LLM-based compression methods were able to achieve a much lower \textit{compression rate} in comparison to previous compression methods. However, if the LLM model size (that can range between several to hundreds of gigabytes) is taken into account (also known as the \textit{adjusted compression rate}), LLM-based compression methods only seems to make sense in data ranging in the terabyte range. Furthermore, existing experiments have not considered the use of LLM-compression on non-English data as well as byte-stream encoded data.

In this work, we revisit existing LLM-compression algorithms with the goal of reducing the adjusted compression rate as well as exploring additional data encoding performance. We show that we are able to reduce the memory footprint of existing LLM's while maintaining similar compression levels. This allows us to achieve a new state-of-the-art (SOTA) adjusted compression rate of nearly $18\%$ without any additional model training. Furthermore, we show that while LLM's are strong compressors for text-dominant data, their performance on non-text data remain competitive when configured correctly. 

The next few sections of this paper are outlined as follows. Preliminaries are presented in Section \ref{sec:preliminaries}. Related works will be discussed in Section \ref{sec:related_works}. Methods will be listed in Section \ref{sec:methods}. Experimental results will be given in Section \ref{sec:methods}. Finally, conclusion and possible extensions will be discussed in Section \ref{sec:conclusions}.

\section{Preliminaries}
\label{sec:preliminaries}
\paragraph{Tokenization}
Tokenization is the process of converting raw text into a sequence of discrete units, known as tokens. These tokens can represent individual characters, subwords, or entire words, and serve as the basic input units for the neural network. By splitting text into tokens, the model can handle language in manageable segments, facilitating both the training and the generation of coherent outputs.

More formally, consider an input text sequence $\mathbf{x} = (x_1, x_2, \ldots, x_N)$. The tokenization process maps $\mathbf{x}$ into a sequence of tokens $\mathbf{t} = (t_1, t_2, \ldots, t_M)$, where each \(t_i\) is drawn from a predefined vocabulary \(V\). A widely used technique, Byte-Pair Encoding (BPE), merges the most frequently occurring pairs of symbols to produce an efficient subword representation. Once the text is tokenized, these tokens can be embedded into continuous vector representations, which the LLM processes through its layers of attention and transformation to generate meaningful predictions and responses.

\paragraph{LLM Training}
LLM's are often trained in a two stage process before applying them for inference. The first stage is \textit{pre-training} and the second stage is \textit{fine-tuning}. 

During pre-training, the LLM is trained on a large text corpus\footnote{Modern pre-training datasets are measured in the order of billions or trillions of tokens. The texts in the corpus are often scrapped from the internet.} to maximize the log-likelihood of each \textit{token} conditioned on preceding tokens. Formally, let $p_\theta(\mathbf{x})$ denote the LLM and let $\theta$ be the parameters of the model. For a given sequence of tokens $\mathbf{x} = \{x_i\}_{i=1}^{N}$ of length $N$, the log-likelihood $L$ is defined as
\begin{equation}\label{log-like}
    L(\Tilde{\theta}, \mathbf{x}) = \log \left( \prod_{i=1}^N p_{\Tilde{\theta}}(x_i|x_{<i}) \right)
\end{equation}
for the sequence $\mathbf{x}$. To optimize the parameters $\theta$ of the LLM, we maximize the expected log-likelihood 
\begin{equation}\label{exp-log-like-max}
    \theta = \arg\max_{\Tilde{\theta}} \mathbb{E}_{\mathbf{x} \sim p(\mathbf{x})} \left[ L(\Tilde{\theta}, \mathbf{x}) \right]
    = \arg\max_{\Tilde{\theta}} \mathbb{E}_{\mathbf{x} \sim p(\mathbf{x})} \left[ \sum_{i=1}^{N} \log p_{\Tilde{\theta}}(x_i|x_{<i}) \right]
\end{equation}
where $x_i$ is the $i$-th token and $x_{<i}$ are the preceding tokens of $x_i$ for a sample $\mathbf{x}$ in the corpus. For completeness, let $x_0$ be the \textit{begin-of-string} token and $p_\theta(x_0)=1$. 

During fine-tuning, the objective is similar to that of pre-training. The difference is that instead of maximizing the log-likelihood across all tokens, we only maximize the conditional log-likelihood of the output token sequence $\mathbf{y} = \{ y_j\}_{j=1}^M$ conditioned on the input sequence $\mathbf{x}$. A set of input and output text sequences $\{(\mathbf{x}, \mathbf{y})\}$ are provided during training, and $\theta$ is optimized by maximizing the expected log-likelihood\footnote{The optimized parameters $\theta$ during pre-training are optimized for a second time during fine-tuning.} (similar to Equation \ref{exp-log-like-max}). After the LLM has been trained, it learns to generate text sequences $\mathbf{y} \sim p_\theta(\mathbf{x})$ given $\mathbf{x}$ as the input. Finding the optimal $\mathbf{x}$ to obtain a desired $\mathbf{y}$ for a given LLM $p_\theta(x)$ is often referred to as \textit{prompting}. 

It has been proven that maximizing the log-likelihood objective (Equation \ref{exp-log-like-max}) is equivalent to minimizing the expected message length (in bits) of an optimal entropy encoder \cite{shannon}. Hence, the motivation is to use LLM's as a probabilistic model to encode data sequences to achieve lossless compression.

\paragraph{Arithmetic Coding}
Arithmetic coding is a form of entropy encoding used in lossless data compression. Unlike traditional coding techniques that assign fixed codewords to symbols (e.g., Huffman coding), arithmetic coding represents a sequence of symbols as a single fractional number in the interval \([0,1)\). This approach allows for more efficient representation, especially when symbol probabilities are not uniform.

\begin{figure}
    \centering
    \includegraphics[width=\linewidth]{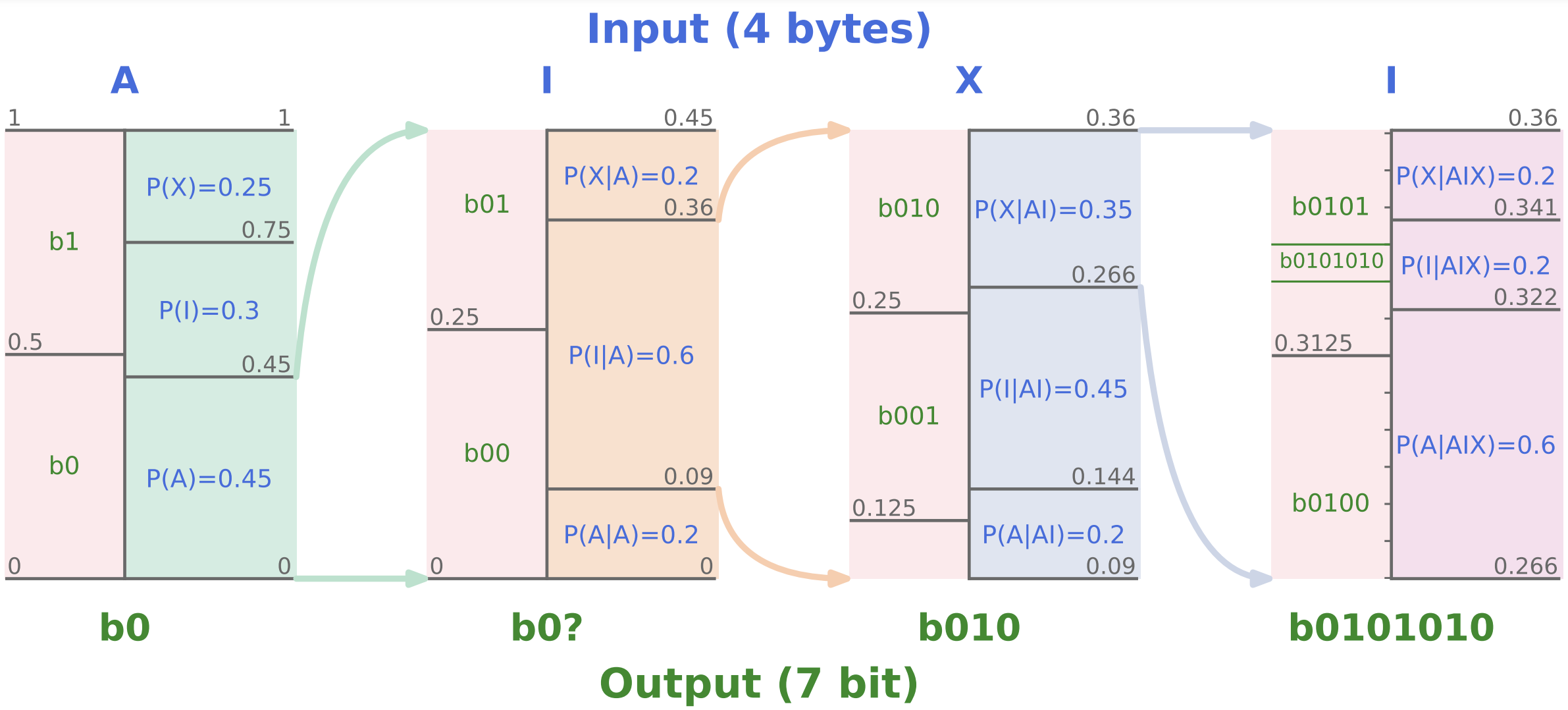}
    \caption{\textbf{Arithmetic coding in action.} (figure from \cite{deletang2024language}) In this example, there are 3 possible symbols: \{X,I,A\}. When encoding a symbol, the probability distribution of all symbols is generated by a probabilistic model (e.g., an LLM). The range for which a valid encoding narrows as the number of encoded symbols increase. The final encoded symbol (e.g., $0101010$) must lie in an interval that represents the original sequence of symbols (e.g., AIXI).}
    \label{fig:arithmetic_coding}
\end{figure}

The encoding process (see Figure \ref{fig:arithmetic_coding}) involves partitioning the interval \([0,1)\) into sub-intervals proportional to the probabilities of each symbol. For a message composed of symbols \(s_1, s_2, \dots, s_n\), the interval is successively narrowed using the probabilities \(P(s)\) of the symbols:

\begin{equation}
    \text{Interval}_0 = [0, 1)
\end{equation}
\begin{equation}
    \text{Interval}_k = [L_k, H_k) = L_{k-1} + (H_{k-1} - L_{k-1}) \times [C(s_k), C(s_k) + P(s_k))
\end{equation}

where \(C(s_k)\) is the cumulative probability of symbol \(s_k\). The final interval \([L_n, H_n)\) uniquely represents the entire message.

Arithmetic coding approaches the theoretical limit of optimal coding efficiency defined by the source entropy $H$:
\begin{equation}
    H = -\sum_{i} P(s_i) \log_2 P(s_i)
\end{equation}

This makes it highly effective for compressing data with known probability distributions, achieving compression rates close to the entropy bound.

\section{Related Works}
\label{sec:related_works}

\paragraph{Arithmetic Coding with LLM}
\label{arithmetic_coding_with_llm}
In the work of \cite{deletang2024language}, the authors explored the use of LLM's and arithmetic coding to compress text, audio, and image data. The first category of LLM's they considered were transformer models \cite{transformers} trained from scratch ranging in the $200$k, $800$k, and $3.2$M parameter range. These LLM's were trained on the enwik7 and enwik8 text-only datasets \cite{hutter2005universal}. The second category of LLM's are pre-trained LLM's from the Chinchilla \cite{chinchilla} family and Llama \cite{Touvron2023Llama2O}. These pre-trained models were trained on significantly larger datasets, and their model sizes range from 7 billion to 70 billion parameters. Evaluation was performed on $1$GB datasets from the enwik9 text dataset\cite{hutter2006prize}, ImageNet \cite{russakovsky2015imagenet}, and LibriSpeech \cite{panayotov2015librispeech} datasets. 

Their results demonstrate that in terms of \textit{raw compression rate} (compressed size $/$ raw size), the LLM-based methods significantly out-perform standard compression algorithms by a large margin. However, when compressing image and audio data, transformers trained on the enwik8 dataset were not able to achieve compression (compression rate $> 100\%$). In contrast, Llama2 \cite{Touvron2023Llama2O} and Chinchilla \cite{chinchilla} models (pre-trained on large scale datasets) were able to achieve a compression rate on image and audio data that surpasses even PNG \cite{boutell1997png} and FLAC \cite{coalson2008flac} compression standards. This is surprising since these models were supposedly trained on text-only datasets.

However, this paper also highlights a disadvantage of LLM-based compression methods. LLM's seem to lose their competitive edge when considering the adjusted compression rate. Their compression abilities are offset by the model weights themselves ($>3$GB for the $1$-billion parameter model). Furthermore, the Llama 2 \cite{Touvron2023Llama2O} and Chinchilla \cite{chinchilla} models that the authors explored are all limited to $\leq 2048$ tokens due to the way these LLM's are trained. In classical compression algorithms, a larger context length allows the compressor to exploit sequential dependencies in the data to achieve a lower compression rate. Due to the limitations in the LLM's context length, the authors have only explored sequences of under $2048$ tokens. This limits the amount of data the LLM is able to process, and it may cause inefficiencies in reducing the compression rate. 

In the work \cite{valmeekam2023llmzip}, the authors explored several approaches of using LLM's in combination with lossless compression algorithms. In addition to arithmetic coding (which matches the same method described by \cite{deletang2024language}), the authors introduce Token Rank Compression (TRC) and Token-by-Token Compression (TTC). We breifly explain their methodology.

\paragraph{Token Rank Compression (TRC)}
Consider the case where we are processing token $x_i$. We condition the LLM on previous tokens $x_{<i}$, and let the LLM predict a probability distribution vector $\Vec{q_i} = [ q_i(1), q_i(2), ..., q_i(D) ]$ across all $D$ tokens in the vocabulary. We sort $\Vec{q_i}$ in a descending order, and record the rank $r_i$ for which the probability corresponding to $x_i$ sits in the sorted $\Vec{q_i}$ (equivalent to the index). The recorded sequence $\{ r_i\}_{i=1}^N$ is then compressed using a lossless compression algorithm (the authors considered the use of zlib). In an ideal setting, the rank of $x_i$ would be at $0$. In reality, the sequence $\{r_i \}$ would fluctuate in small non-negative integers. 

\paragraph{Token-By-Token Compression (TTC)}
Consider again the case where we are processing token $x_i$. Given previous tokens $x_{<i}$, we predict the probability distribution vector $\Vec{q_i} = [ q_i(1), q_i(2), ..., q_i(D) ]$ for token $x_i$. Next, we construct a prefix-free code of length $l_i=\lceil \log_2 \frac{1}{q_i(x_i)}\rceil $, which satisfies the Kraft inequality. A natural choice for such a prefix-free code is the Huffman code \cite{huffman_coding}, but any prefix-free code would suffice. Following such a scheme, there would be a time-varying code-book developed for every token processed.

The results from their experiments reveal that LLM with arithmetic coding yields the best result among the three methods, with TTC yielding similar but slightly sub-optimal performance. The authors have also experimented with compressing out-of-distribution text data. Their experimental results support the claim that a trained LLM is able to compress out-of-distribution text data just as effectively.

\paragraph{Tree Building with $n$-gram Models}
In the work \cite{el_daher_connor_2006}, the authors considered the use of trigram, bigram, and unigram models as language models. Their algorithm is split between a tree-building and compression-decompression phase. We briefly describe their approach in the following.

During the tree building phase, the trigram model is built by modeling the probability $p(x_3|x_1x_2)$ of each subsequent word $x_3$ following the preceding two words $x_1x_2$. This means that for every bigram $x_1x_2$, the language model would have probability distribution over $x_3$. A special 'UNK' token is used to denote any unknown words not in the vocabulary of the trigram model. The probability distribution also includes the proability of the 'UNK' token. 

Constructing the bigram and unigram models follows the same procedure as the trigram model in computing the distribution over $p(x_2|x_1)$ and $p(x_1)$. For cases where the single word is not present in the unigram vocabulary, a character compression model splits the characters inside the word, and computes individual character-level probabilities.

For each trigram, build a Huffman tree using the probability distribution $p(x_3|x_1x_2) \; \forall x_1x_2$ in the trigram vocabulary. Similarly, build a Huffman tree using the probability distribution $p(x_2|x_1) \; \forall x_1$ in the bigram vocabulary. Lastly, build a Huffman tree using the probability distribution $p(x_1)$ of the unigram vocabulary. Note that all trigram, bigram, and unigram all contain the 'UNK' token. 

To compress the data, we go through each word (there are 2 start tokens to begin with). We look up the current word using the trigram model. If the word exists within the trigram vocabulary, we encode it using its corresponding Huffman code, and move to the next word. If the word is not found, we write the code for the 'UNK' token, and revert to the bigram model. Similarly, if the word exists, we record its Huffman code. If the word does not exist, record the 'UNK' Huffman code and revert to the unigram model. We continue a similar procedure on the unigram model, and if the word does not exist, we encode the word at the character level using the character model. This encoding process is repeated on all words in the given text, and the end result is a concatenation of all encoded Huffman codes.

To decompress an encoded sequence, we begin a lookup using the Huffman tree corresponding to the initial two start tokens. Decode the sequence using the corresponding trigram Huffman tree. If the decoded token is the 'UNK' token, continue to decode the remaining sequence using the bigram Huffman tree. If the resulting sequence is the 'UNK' token, continue the decoding process using the unigram tree. If the resulting sequence is the 'UNK', then continue the decoding using the character level Huffman tree. If during this decoding process, the decoded sequence is not the 'UNK' token, revert the tree back to the trigram case after the last encoding, and use the last two previously decoded token to select the relevant trigram Huffman tree. Continue this process until the encoded sequence has been decoded.

An interesting choice in this algorithm is the decision to limit the $n$-gram models to just a trigram. The authors explained that although a longer sequence will allow stronger dependencies between words, longer $n$-grams will suffer from data sparsity. That is why they limit their design to a trigram at maximum.

The authors compared their algorithm to the BZip2 and GZip compressors. Their proposed approach closely matches that of GZip without any optimizations. In addition, they experimented with adding more training data to the tree-building process. They observed a continual reduction in compression rate as more training data is added during tree-building. By adding more training data, it reduces the probability of encountering an 'UNK' token, which causes a prolonged sequence to be encoded.

\paragraph{Image, Video, and Audio Compression}
In this work \cite{li2024understandingcompression}, the authors decided upon image-GPT \cite{chen2020generative} as the probabilistic model for image compression. This is a different probabilistic model for the image domain in comparison to the text-prominent LLM proposed by \cite{deletang2024language}. Image-GPT is a large-scale vision model that has been trained on a wide range of image data. It generates images in an auto-regressive manner by predicting the probability of each pixel in a sequence. This sequence-by-sequence generation properties allows the authors to use arithmetic coding as the compression algorithm. 

Since images are essentially 2D matrices (if we discard the channel dimension), compressing an image using Image-GPT requires resizing the 2D matrix into a one-dimensional sequence of pixels. Unlike dealing with text sequences, pixels values ranges from $[0, 255]$, and hence, the probabilities can be trivially obtained. Due to the limited context window of the Image-GPT, the authors decided to divide the sequence into non-overlapping segments to fit within the Image-GPT's context window. The compressed segments are then concatenated to form the final probability, which is encoded using arithmetic coding.

Video compression is done in a similar manner to image compression. Unlike classical methods that exploit inter-frame information for compression, the authors did not exploit such dependencies. They were concerned that video containing drastic motion changes would not benefit from inter-frame dependencies. Their second concern is that they have found that exploiting inter-frame dependence for slow-changing inter-frame videos did not really help the overall compression performance. Hence, the authors reverted to single frame compression as in image compression.



On image data, LMCompress was able to achieve a much higher compression rate (higher is better) compared to classical methods. When compared to Chinchilla 7B and 70B from \cite{deletang2024language}, LMCompress was able to benifit from the image auto-regressive properties of the Image-GPT, resulting in over 2 times higher compression rate than LLM based methods. Video compression also benifited with the use of Image-GPT, as the compression rate was over 20\% higher than common lossless video compressors like FFV1, H.264, and H.265. Audio compressino also benifited with the custom-trained audio LLM. In comparison, standard language-centric LLM's such as Llama \cite{Touvron2023Llama2O} and Chinchilla \cite{chinchilla} achieve an average compression rate of between $4.24 - 4.76$. LMCompress was able to achieve a compression rate of $6.07$. This result emphasize the importance of having the probabilistic model that can model the data sequence well.

\section{Methods}
\label{sec:methods}

The goal of this work is to achieve an adjusted compression rate using LLM-based compression that is able to maintain competitiveness with existing general purpose compressors. As demonstrated in \cite{valmeekam2023llmzip, deletang2024language}, LLM-based probabilistic modeling with arithmetic coding achieves near theoretical optimal compression. Hence, we consider the use of arithmetic coding, and we explore different methods to enhance or maintain the model performance while maintaining or reducing model size. We discuss several possibilities in the following sections.

\subsection{Context Length Extension} 
\label{methods:context_length_extension}
One method to improve LLM performance is to increase the context length for which a sequence is to be processed. The context length effectively determines the amount of bytes a model is able to compress at a time. In the works of \cite{valmeekam2023llmzip, deletang2024language}, it has been shown that increasing the context length of a model would often result in an improved compression rate. This is because the model is able to exploit more inter-token dependencies, allowing for better predictability of each subsequent token.

The two main challenges with increasing an LLM's context lengths are the model's inherent support for long contexts and its memory usage. Most modern LLM's are trained using a form of Rotary Positioning Embedding (RoPE) \cite{ropeextension} with a pre-define context length. RoPE embeddings encode the absolute and relative position of each input token's embedding to allow the LLM to understand the order of the input tokens. If a sequence that exceeds the context length of a given LLM is provided, the LLM would often struggle with encoding the positional information of the input tokens, which causes the LLM to deteriorate in performance.

Some methods \cite{peng2024yarn, Chen2023ExtendingCW} have been proposed to increase the context length of such positional embeddings to allow the LLM to extend beyond their original context lengths. However, these methods typically involve additional post-training, and require long context data trained on designated pools of GPU clusters. Even if the LLM has been trained with large context length support, running the LLM with a large context length for inference purposes would also require significant GPU memory usage. This is because the memory usage increases quadratically following increases in the context length due to the self-attention mechanism within transformers \cite{transformers}. In this work, we consider LLM models that have long context length support, and we evaluate the improvements from to longer context lengths. 

\subsection{Reducing Model Size}
\label{methods:reducing_model_size}
The main challenge in reducing the adjusted compression rate comes from the large model size of such LLM's. Most modern LLM's are trained and stored in Float16 or BFloat16 precision, which takes up 2 bytes per parameter. To reduce the model size of existing LLM's, several methods \cite{frantar-gptq, lin2023awq, badri2023hqq} have been proposed to quantize LLM model weights to either Int8, Int4, or even Int2 precision. The goal of LLM model compression is to maintain the same performance as its full-precision model while storing the model at a fraction of the original size. 

Quantization methods can occur during training (i.e., quantization-aware training) or after training (i.e., post-training quantization). Since we do not consider re-training an LLM from scratch in this work, we focus on post-training quantization as our main focus. The two main types of existing post-training quantization can be divided into calibration-based and calibration-free quantization. Specifically, we consider a candidate from each quantization type as part of our work.

\subsubsection{GPTQ}
\label{methods:gptq}
GPTQ \cite{frantar-gptq} is a one-shot calibration-based weight quantization algorithm. For each layer $l$ of a given LLM model, the GPTQ algorithm's objective is to find a quantized weight matrix $\mathbf{\widehat{W}}_l$ of the full-precision weight matrix $\mathbf{W}_l$. Since the weights matrix $\mathbf{\widehat{W}}_l$ of each layer is multiplied by an input $\mathbf{X}$ during the forward-pass, this quantization should ensure that the output activations using the original weight matrix $\mathbf{W}_l \mathbf{X}$ and the quantized weight matrix $\mathbf{\widehat{W}}_l \mathbf{X}$ remain nearly identical. Formally, the objective can be defined as

\begin{equation}
    \label{eq:layerwise-quantization}
    \text{argmin}_{\mathbf{\widehat{W}}_l} \, ||\mathbf{W}_l \mathbf{X} - \mathbf{\widehat{W}}_l \mathbf{X}||_2^2
\end{equation}

for each layer $l$ in the LLM.

The GPTQ algorithm builds upon the OBQ \cite{obqquantization} algorithm, which performs greedy quantization of weight matrix items one weight after another. For smaller models (several million parameters), the quantization process could take a few hours. This would become a big bottleneck when trying to quantize much larger billion-parameter LLM's. 

GPTQ performs a column-wise quantization by designating blocks of size $B$, which consists of $B$ columns. In each block, the $j$-th column $\mathbf{W}_{:j}$ is quantized to $\mathbf{Q}_{:j}$. Th quantization error $\mathbf{E}$ is then computed, which is used to update the values of the weight matrix corresponding to the block. The weights of the remaining un-quantized weight matrix is updated only after a block has been quantized. The full algorithm is given in Algorithm \ref{alg:gptq}, and the full details are discussed in \cite{frantar-gptq}.

\newlength{\commentindent}
\setlength{\commentindent}{.45\textwidth}
\makeatletter
\renewcommand{\algorithmiccomment}[1]{\unskip\hfill\makebox[\commentindent][l]{\textit{//~#1}}\par}
\LetLtxMacro{\oldalgorithmic}{\algorithmic}
\renewcommand{\algorithmic}[1][0]{%
  \oldalgorithmic[#1]%
  \renewcommand{\ALC@com}[1]{%
    \ifnum\pdfstrcmp{##1}{default}=0\else\algorithmiccomment{##1}\fi}%
}
\makeatother

\begin{algorithm}[]
    \centering
    \caption{Quantize $\mathbf{W}$ given inverse Hessian $\mathbf{H}^{-1} = (2 \mathbf{X} \mathbf{X}^\top + \lambda \mathbf{I})^{-1}$ and blocksize $B$.}
    \small
    \label{alg:gptq}
    \begin{algorithmic}
        \STATE $\mathbf{Q} \gets \mathbf{0}_{d_\text{row} \times d_\text{col}}$ \quad \COMMENT{quantized output}
        \STATE $\mathbf{E} \gets \mathbf{0}_{d_\text{row} \times B}$ \quad \COMMENT{block quantization errors}
        \STATE $\mathbf{H}^{-1} \gets \text{Cholesky}
        (\mathbf{H}^{-1})^\top$ \COMMENT{Hessian inverse information}
        \FOR {$i = 0, B, 2B, \dots$}
            \FOR {$j = i, \dots, i + B - 1$}
                \STATE $\mathbf{Q}_{:, j} \gets \text{quant}(\mathbf{W}_{:, j})$ \quad \COMMENT{quantize column}
                \STATE $\mathbf{E}_{:, j - i} \gets (\mathbf{W}_{:, j} - \mathbf{Q}_{:, j}) \, / \, [\mathbf{H}^{-1}]_{jj}$ \COMMENT{quantization error}
                \STATE $\mathbf{W}_{:, j:(i + B)} \gets \mathbf{W}_{:, j:(i + B)} - \mathbf{E}_{:, j - i} \cdot \mathbf{H}^{-1}_{j, j:(i + B)}$ \COMMENT{update weights in block}
            \ENDFOR
            \STATE $\mathbf{W}_{:, (i + B):} \gets \mathbf{W}_{:, (i + B):} - \mathbf{E} \cdot \mathbf{H}^{-1}_{i:(i + B), (i + B):}$ \COMMENT{update all remaining weights}
        \ENDFOR
    \end{algorithmic}
\end{algorithm}

The main issues with calibration-based quantization algorithms are that calibration samples are required and the quantization compute time is usually lengthy. Ideally, the calibration data should originate from the training samples, but in the case of pre-trained models, such data are usually unavailable. Furthermore, the choice of the calibration data may cause bias during the calibration process. As described in GPTQ \cite{frantar-gptq} quantization, multiple forward passes are required to quantize each weight matrix. This introduces additional compute and delays during model development.

\subsubsection{HQQ}
\label{methods:hqq}
Half-Quadratic Quantization (HQQ) \cite{badri2023hqq} is a calibration-free quantization algorithm. The focus of HQQ is to align the weights $\mathbf{W}$ rather than aligning the model activations $\mathbf{WX}$. In essence, the goal is to find a quanzation operator $\mathbf{Q}_{z,s}(\mathbf{W})$ that satisfies 

\begin{equation}
    \label{eq:layerwise-quantization}
    \text{argmin}_{z, s} \, \varphi( \mathbf{W} - \mathbf{Q}_{z,s}^{-1}(\mathbf{Q}_{z,s}(\mathbf{W})).    
\end{equation}

Here, $\varphi()$ is a sparsity-promoting loss that better captures the outlier errors often located in the heavy tails of the error distribution. The quantization operator $\mathbf{Q}_{z,s}$ and its inverse (de-quantization operator) $\mathbf{Q}^{-1}_{z,s}$ is defined as

\begin{equation}
\label{eq:hqq_quant}
    \mathbf{Q}_{z,s}(\mathbf{W}) = \widehat{\mathbf{W}} = \text{round}(\mathbf{W}/s + z)
\end{equation}

\begin{equation}
\label{eq:hqq_dequant}
    \mathbf{Q}^{-1}_{z,s}(\widehat{\mathbf{W}}) = s(\widehat{\mathbf{W}} - z)
\end{equation}.

To solve this problem, the authors proposed the use of a Half-Quadratic Solver \cite{half_quad_solver}. The scaling term $s = 1$ is fixed, and a new variable $\mathbf{W}_e$ is introduced to optimize for

\begin{equation}
\label{eq:hqq_opt_obj}
    \text{argmin}_{z, \mathbf{W}_e} \, \varphi( \mathbf{W_e}) + \frac{\beta}{2} || \mathbf{W}_e - (\mathbf{W} - \mathbf{Q}_z^{-1}(   \mathbf{Q}_z(\mathbf{W})))||_2^2.
\end{equation}.

Equation \ref{eq:hqq_opt_obj} can be solved by alternate optimization between Equations \ref{eq:hqq_sol1} and  \ref{eq:hqq_sol2}. Parameters $\alpha$ and $\beta$ are positive hyper-parameters that should be updated after each alternate step (see Equation \ref{eq:hqq_sol3}).

\begin{equation}
\label{eq:hqq_sol1}
    \mathbf{W}_e^{(t+1)} \leftarrow \text{argmin}_{\mathbf{W}_e} \, \varphi( \mathbf{W_e}) + \frac{\beta^{(t)}}{2} || \mathbf{W}_e - (\mathbf{W} - \mathbf{Q}_z^{-1}(   \mathbf{Q}_z(\mathbf{W})))||_2^2.
\end{equation}.
\begin{equation}
\label{eq:hqq_sol2}
    z^{(t+1)} \leftarrow \text{argmin}_z \,\, \frac{1}{2} || \mathbf{Q}_z^{-1}(   \mathbf{Q}_z(\mathbf{W})) - (\mathbf{W} - \mathbf{W}_e^{(t+1)}||_2^2
\end{equation}.
\begin{equation}
\label{eq:hqq_sol3}
    \beta^{(t+1)} \leftarrow \alpha \beta^{(t)}
\end{equation}

Unlike gradient descent, optimizing for $z$ using the Half-Quadratic Solver can be completed in a few iterations without any forward passes on the model. HQQ can be applied to the weight matrices in an LLM to obtain calibration-free quantization. 

\subsection{Tokenization Methods}
As noted by \cite{valmeekam2023llmzip}, tokenization in language modeling is essentially a form of lossless compression. LLM tokenizers are typically trained separately  on a given text corpus prior to using them to pre-train an LLM. During tokenizer training, the tokenizer's vocabulary set is determined by accounting for the most frequent texts in the corpus. After tokenizer training has completed, the vocabulary set is kept fixed, and only the embeddings of the vocabulary and the model weights are trained during LLM pre-training and fine-tuning.

Although LLM's seem to handle text compression well, their compression performance on non-text data is yet to be understood. In this work, we consider data types in the form of a byte stream. Since all data encodings (e.g., PDF's, PPT, Excel) are encoded in bytes in their raw format, we consider the case where only the byte stream is available, and we would like to know whether LLM's are able to compress such data. Next, we discuss some possible ways to perform compression on byte streams. 

\renewcommand{\arraystretch}{1.2}

\begin{table}[h!]
\label{tab:utf8_table}
\centering

\begin{tabularx}{\textwidth}{l l X X X X}
\toprule
\textbf{First code point} & \textbf{Last code point} & \textbf{Byte 1} & \textbf{Byte 2} & \textbf{Byte 3} & \textbf{Byte 4} \\
\midrule
U+0000   & U+007F   & \texttt{0yyyyzzzz} &                 &                 &                 \\
U+0080   & U+07FF   & \texttt{110xxxyy} & \texttt{10yyzzzz} &                 &                 \\
U+0800   & U+FFFF   & \texttt{1110wwww} & \texttt{10xxxxyy}  & \texttt{10yyzzzz} &                 \\
U+010000 & U+10FFFF & \texttt{11110uvv}  & \texttt{10vvwwww}  & \texttt{10xxxxyy} & \texttt{10yyzzzz} \\
\bottomrule
\end{tabularx}
\caption{\textbf{Code point $\leftrightarrow$ UTF-8 conversion.} The value of the code point can be encoded using $1-4$ bytes. The code point in hexadecimal representation is \texttt{U+uvwxyz}}

\end{table}

\subsubsection{Treating Byte Streams as Text}
\label{met:tok:byte_as_text}
This is perhaps the most straightforward method. This strategy treats the byte stream as a string by converting each byte into their UTF-8 character representation. The character are then concatenated to form one big string. The string is then tokenized using an LLM tokenizer, and we perform LLM compression (see Section \ref{arithmetic_coding_with_llm}). 

\subsubsection{Treating Byte Streams as Integers}
\label{met:tok:byte_as_int}
This strategy treats the byte stream as values represented by their byte representations. We consider values between $0-255$, and we map the bytes to their integer representations in that range. Next, we configure the LLM tokenizer such that we only consider the tokens representing the values in $\{0, 1, 2, ..., 255\}$. During tokenization, we map each integer from the byte array into their corresponding token from the tokenizer, and perform LLM compression using these mapped tokens.

\begin{figure}[t]
    \centering
    \includegraphics[width=\linewidth]{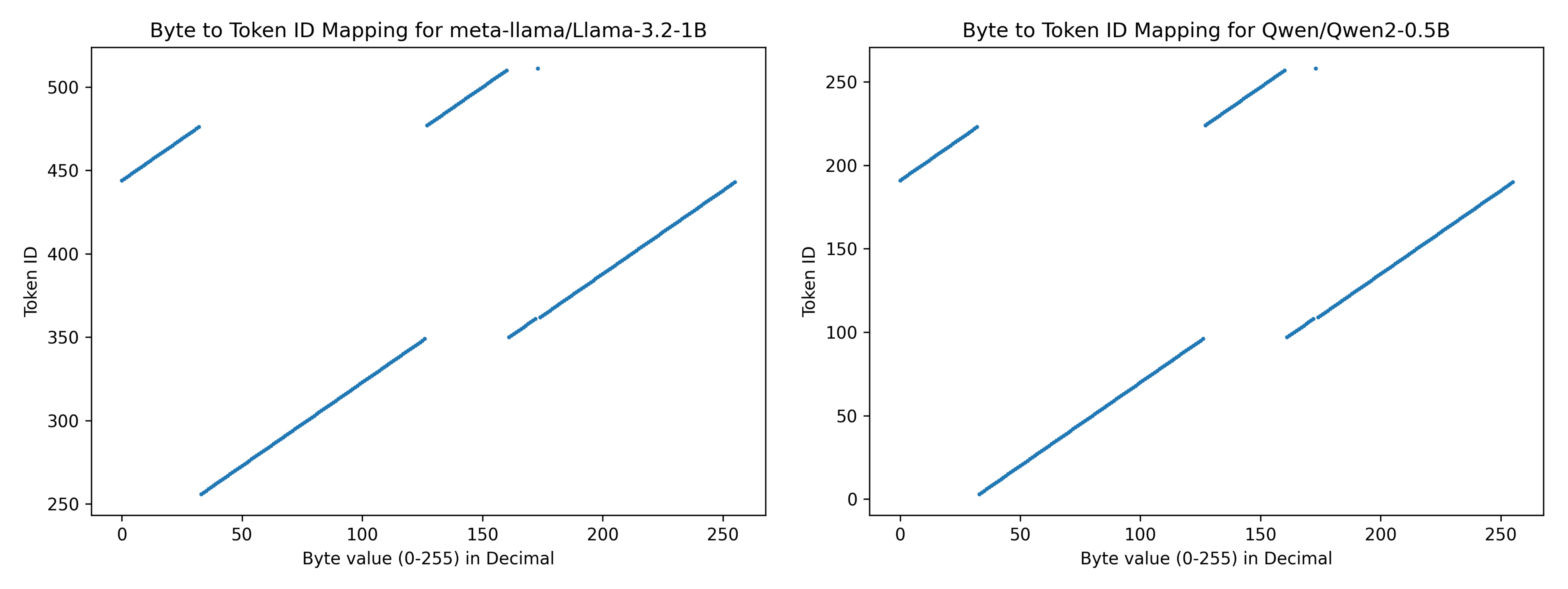}
    \caption{\textbf{Byte Mapping for Llama 3 and Qwen Model.} On the left is mapped byte tokens for the Llama 3 \cite{Touvron2023Llama2O} model. On the right is the mapped byte tokens for the Qwen \cite{qwen} model. Notice how the two mappings are seemingly identical, except that their mapped tokens are shifted by approximately $200$. This is most likely due to the fact that both tokenizers are based on Byte-Pair Encoding. The different in token value shift is due to different configurations of the two tokenizers.}
    \label{fig:byte_mapping}
\end{figure}

\subsubsection{Treating Byte Streams as Bytes}
\label{met:tok:byte_as_byte}
This method is similar to treating the byte stream as integers. However, instead of mapping the integer values to their corresponding token in the tokenizer, we map the integers to their corresponding bytes in the tokenizer. Recall that most existing LLM's are based on Byte-Pair Encoding. This means that in the case where a word does not exist in the tokenizer's vocabulary, the tokenizer will break this word into its byte representation, which is represented by more than one token. However, the challenge is finding the tokens for which each (actual) byte value corresponds to. Here, we briefly explain how we found the mapping from each byte to its corresponding token in the tokenizer. 

We initialized a tokenizer from an existing tokenizer (to ensure identical configurations), and we removed all tokens from the the tokenizer's vocabulary. This means that all texts tokenized by this empty tokenizer is now all in its byte tokens. Next, we followed the UTF-8 encoding table (see Table \ref{tab:utf8_table}) to iterate through valid byte combination values. For each valid combination, we tokenized the corresponding UTF-8 decoded string using the empty tokenizer to obtain the tokenizer's byte token. Since the some values in cannot be decoded following the UTF-8 coding schema (e.g., 0xCO, 0xC1, 0xF5-0xFF), we first map all values values. We then interpolate these missing values from their neighboring values. The resulting mappings can be seen on Figure \ref{fig:byte_mapping}.

Using this mapping, we map each byte into their corresponding token, and we send these tokens into the LLM for compression. 

\section{Experiments}
In this section, we explore the different methods we have tried to achieve a reduced \textit{adjusted compression rate}. We then revisit LLM compression on tasks involving non-English data and evaluate the performance of LLM compression compared to general purpose compressors. We begin by introducing our setup. Experimental results will be presented, which is followed by our analysis.

\subsection{Setup}
\subsubsection{Models}
The focus of this work is on the compression performance of pre-trained LLM's. Hence, we do not consider methods that require additional training. The models presented in \cite{deletang2024language} are the 1B, 7B, and 70B models of the Chichilla \cite{chinchilla} family and the 7B model of the Llama 2 \cite{Touvron2023Llama2O} family. Since the release of Llama 2, Meta has released Llama 3 models of smaller sizes. In particular, we consider the Llama 3.2 1B model released by Meta. In comparison to Llama 2, Llama 3 models have extended the tokenizer vocabulary from the origianl $32$k to a staggering $128$k. This extended size allows for more token representations, indicating that less texts would be converted to their raw byte representation during tokenization. However, this increase in tokenizer vocabulary also increases the model parameter (in Llama 3.2 1B, each token is mapped to a unique vector of dimension $2048$). 

Another family of models we consider is the Qwen 2 \cite{qwen} series. These model were trained by the Qwen team from Alibaba, and the model they released are in the sizes of $0.5$B, $1$B, $3$B, $7$B, $14$B, $32$B, and $72$B parameters. In the interest of model size reduction, we consider their smallest model at the $0.5$B scale. In addition to the Qwen series of models, we consider the SmolLM \cite{smollm} family of models developed by a team from Huggingface. The SmolLM series of models come at sizes of $150$M, $360$M, and $1.7$B parameters trained on fully synthetic data generated by other LLM's. We consider the model of sizes $150$M and $360$M parameters in the interest of smaller model sizes. 

\subsubsection{Datasets}
To evaluate the compression performance of existing LLM's on English and non-English data, we consider text datasets of $1$GB sizes from English (enwik9) \cite{hutter2006prize}, Spanish \cite{ramitha_spanish_legal_data_2}, French \cite{huggingface2024croissant}, Russian \cite{RussianNLP_Mixed_Summarization_Dataset}, Thai \cite{thai_dataset}, Japanese \cite{japanese_dataset}, Arabic \cite{arabic_text}, and Traditional Chinese (Taiwan) \cite{taiwan_excellence} languages. We include code code data \cite{code_parrot} as part of our evaluation for language data. For byte stream data, we consider PDF files \cite{pdf_data}.

\subsubsection{Metrics}
\label{exp:setup:metrics}
To evaluate the effectiveness of each compression algorithm, we consider the \textit{raw} compression rate $\gamma_r$ and \textit{adjusted} compression rate $\gamma_a$. The raw compression rate $\gamma_r$ is defined as 

\begin{equation}
    \gamma_r = (\text{compressed size in bytes})/ (\text{original size in bytes}).
\end{equation}

In contrast, the adjusted compression rate $\gamma_a$ is defined as

\begin{equation}
    \gamma_a = (\text{compressed size in bytes} + \text{model weight in bytes})/(\text{original size in bytes}). 
\end{equation}

The closer the compression rate $\gamma \rightarrow 0$ approaches zero, the better the compression performance. If the compression rate $\gamma \rightarrow 1$ approaches one, it indicates that no compression is occurring. If the compression rate $\gamma > 1$, it indicates that the compression has failed to compress the data. This occurs when the probabilistic model fails to predict the data sequence.

\subsubsection{Compute}
We run all our experiments on a NVIDIA RTX 4090 24GB GPU running Python $3.10$ with PyTorch $2.5.1$. Models are loaded from Huggingface distributed by their respective developers.

\subsection{Compression Rate Reduction}
The goal of this set of experiments is to understand the most effective way to reduce the compression rate. We follow the methods introduced in Sections \ref{methods:context_length_extension} and \ref{methods:reducing_model_size} as reference. 

\subsubsection{Context Length Extension}
In this experiment, we explore the possibilities to reduce the compression rate $\gamma$ by increasing the context length of the LLM. We consider the performance of each LLM in their BFloat16 representation, and we evaluate their performance on the enwik9 dataset. Both Llama-3.1-1B and Qwen-2-0.5B natively support context lengths of up to $128$k tokens. In contrast, SmolLM support context lengths of up to $2048$ tokens. Due to the limit in compute resources, we explore a context length of at most $4096$ tokens for our experiments. 

\begin{figure}
    \centering
    \includegraphics[width=0.8\linewidth]{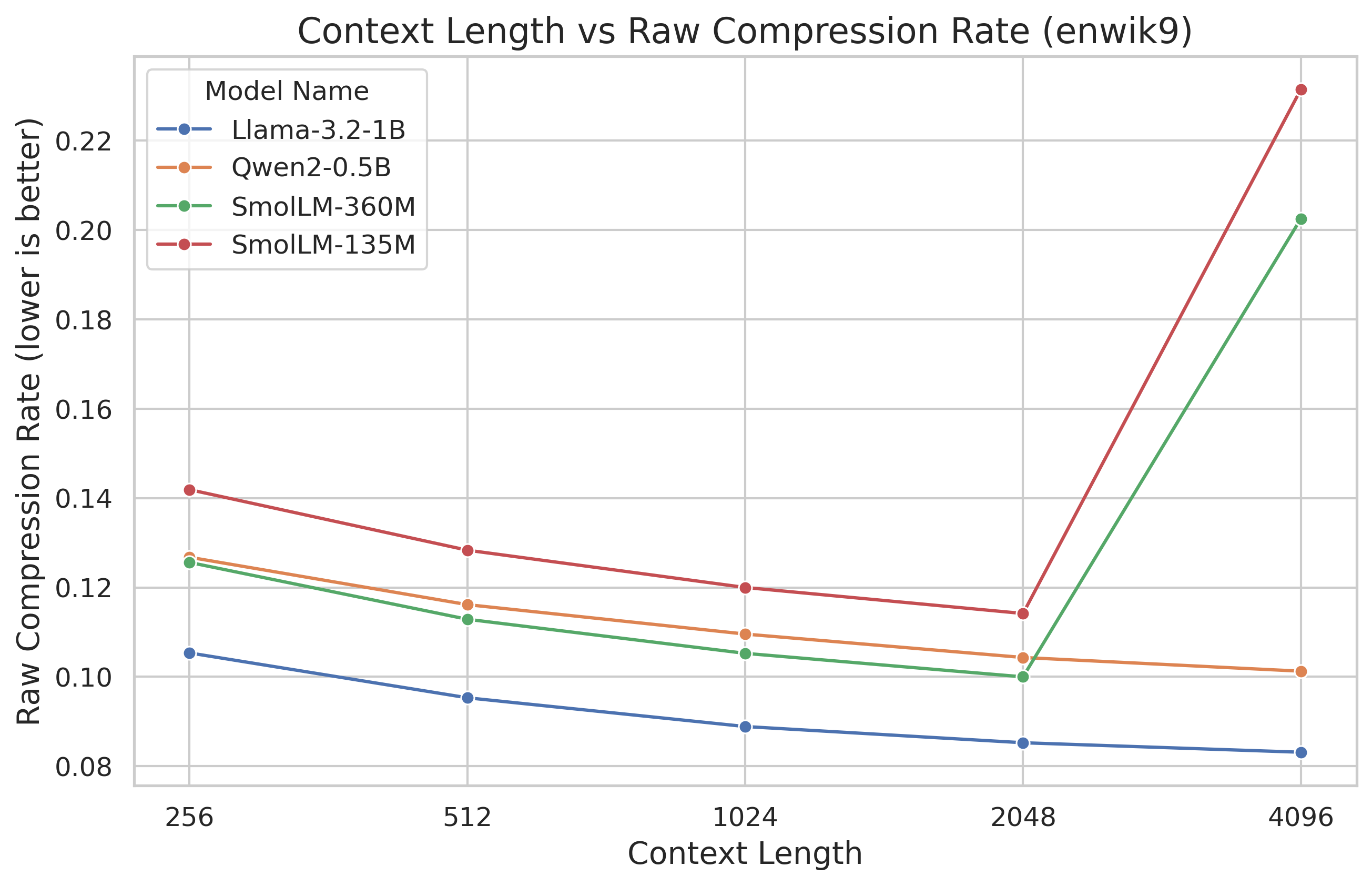}
    \caption{\textbf{Context Length and Compression Rate.} Shown above is the raw compression rate $\gamma_r$ as a function of the context length of an LLM. Notice the steady decrease in $\gamma_r$ as the context length increases. Since Llama-3.2-1B and Qwen2-0.5B both naively support context lengths of up to $128$k tokens, their compression ration is expected to improve for even longer context lengths. Since SmolLM models only support up to $2048$ tokens, $\gamma_r$ increases when we exceed this limit.}
    \label{fig:exp:context_length_ext}
\end{figure}

As we can see in Figure \ref{fig:exp:context_length_ext}, the longer the context length, the more the reduction in the compression rate $\gamma$. However, this reduction only occurs on an incremental scale (i.e., decreasing around $\Delta \gamma_r \sim 0.01$ for every exponential increase in context length). In addition, notice that for the SmolLM models, context lengths exceeding $2048$ yields a sudden increase in compression rate. This is due to the limited context length support of SmolLM models to only under $2048$ tokens. For the remaining of the experiments, we limit the context length to $2048$ tokens as it achieves a good balance between GPU memory and an acceptable compression rate for all models.

\subsubsection{Model Size Compression}

\begin{figure}
    \centering
    \includegraphics[width=\linewidth]{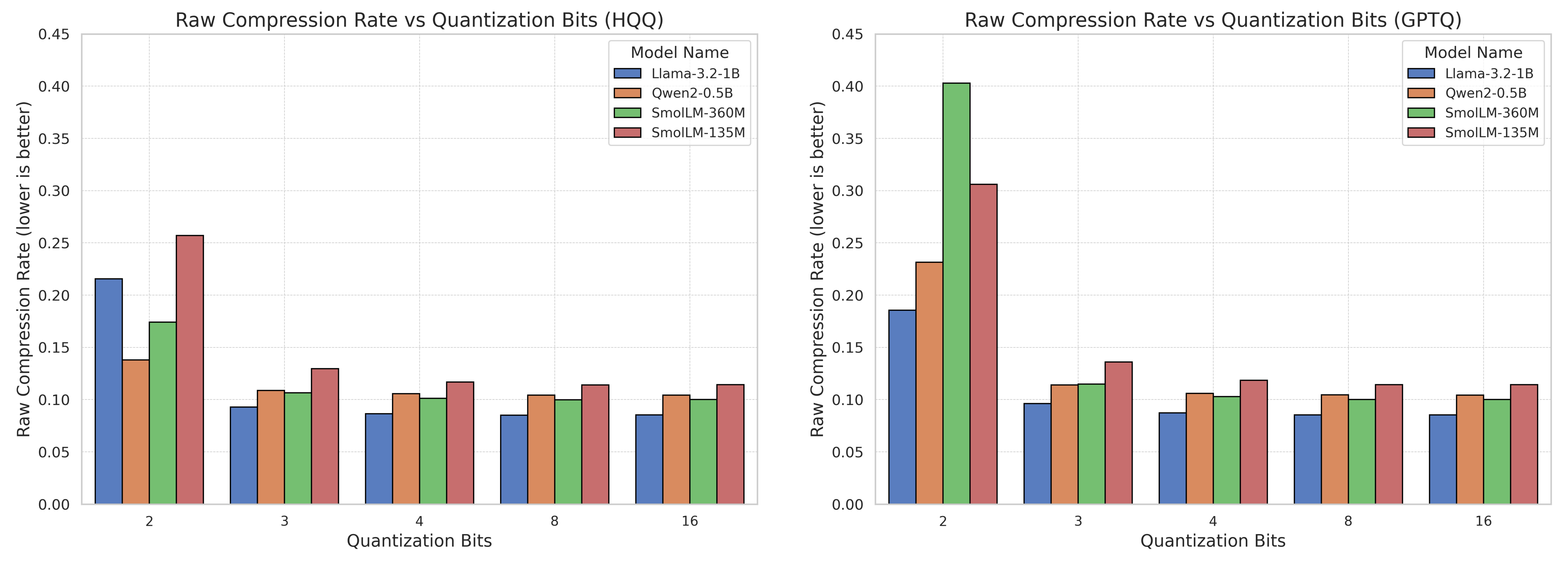}
    \caption{\textbf{Raw Compression Rate Under Different Quantization Schemes.} Show above is the raw compression rate $\gamma_r$ under HQQ (left) and GPTQ (right) quantization. The $16$-bit representation is the original un-quantized representation. Notice how HQQ, a calibration-free quantization algorithm, retains better predictive performance compared to GPTQ.}
    \label{fig:quant:raw_comp_ratio}
\end{figure}

To reduce the model size, we consider the GPTQ\cite{frantar-gptq} and HQQ\cite{badri2023hqq} algorithms introduced in Sections \ref{methods:gptq} and \ref{methods:hqq}. For GPTQ calibration, we sample $128$ text samples from the enwik9 \cite{hutter2006prize} dataset and perform quantization with block size $B=8$. We quantize the model sizes from their original BFloat16 representation to their $8$-bit, $4$-bit, $3$-bit, and $2$-bit representations. For HQQ quantization, we do not perform calibration as it is not needed.

The raw compression rates $\gamma_r$ of the different quantization schemes are presented in Figure \ref{fig:quant:raw_comp_ratio}. From the raw compression rate, notice that the more intense the quantization (i.e., using less bits to represent the original BFloat16 weights), the higher the resulting compression rate. Despite this increasing trend, we are able to compress the model weights from their original $16$-bit representation down to a $3$-bit representation (over $5 \times$ reduction in model size) with a slight drawback in compression rate. This ability to maintain a strong compression rate only deteriorates at the $2$-bit quantization level. Furthermore, this issues is seemingly more prominent in GPTQ compression. 

\begin{figure}
    \centering
    \includegraphics[width=\linewidth]{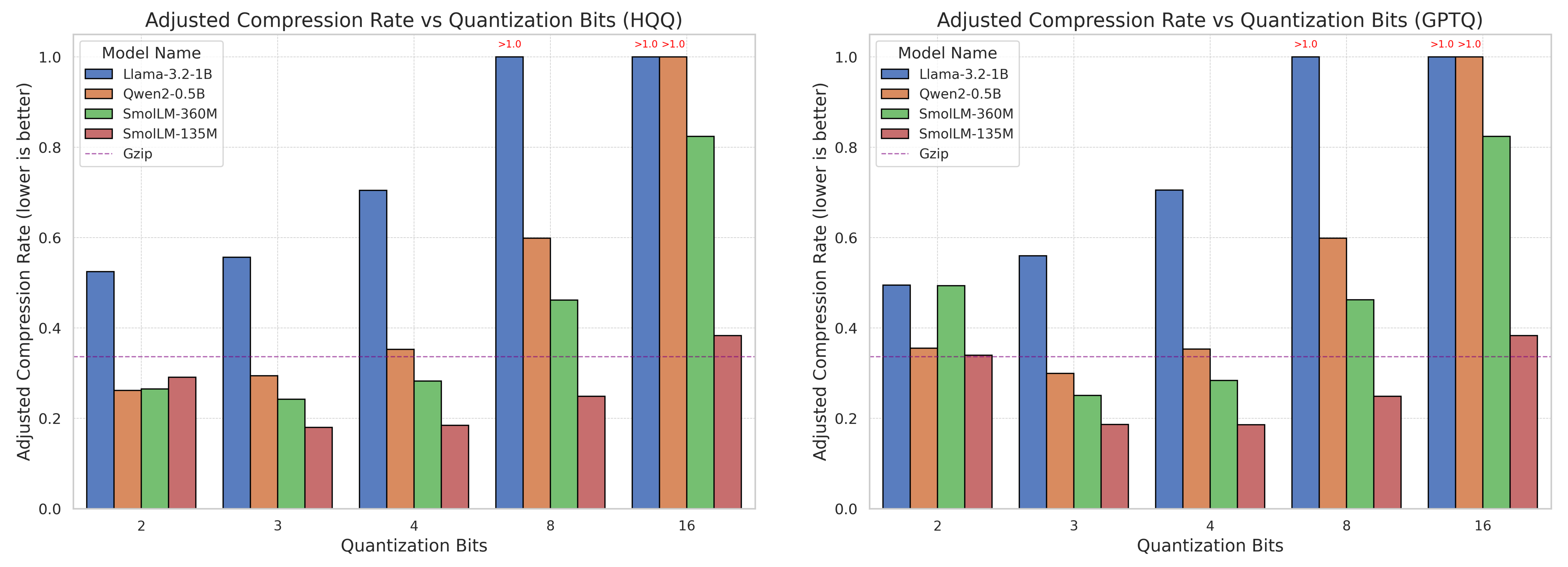}
    \caption{\textbf{Adjusted Compression Rate Under Different Quantization Schemes.} Show above is the adjusted compression rate $\gamma_a$ under HQQ (left) and GPTQ (right) quantization. The $16$-bit representation is the original un-quantized representation. Compression rates exceeding $1.0$ have been labeled and capped. In comparison to the baseline general purpose compressor, LLM's under the $0.5$B parameter range have all exceeded Gzip's compression rate with $3$-bit compression.}
    \label{fig:quant:adj_comp_ratio}
\end{figure}

The adjusted compression rate $\gamma_a$ under different quantization schemes is illustrated in Figure \ref{fig:quant:adj_comp_ratio}. In comparison to the standard general-purpose compressor Gzip, our quantized LLM's are able to maintain a strong competitive advantage at both the $4$-bit and $3$-bit quantization level. Despite the increase in compression rate at $2$-bit quantization, HQQ compression still outperforms Gzip in LLM's under $0.5$B parameters, with the lowest $\gamma_a = 0.18$. This result sets a new state-of-the-art (SOTA) in un-trained LLM compression performance in comparison to \cite{deletang2024language}. The complete results for this experiment is provided in Table \ref{table:quantization}. For the remaining of the experiments, we consider the HQQ quantized models at $3$-bit representation since this configuration offers a good balance between model size and adjusted compression rate.

\begin{table}[]
\centering
\small
\begin{adjustbox}{max width=\textwidth}
\begin{tabular}{l l c S S}
\toprule
\textbf{Model} & \textbf{Quantization} & \textbf{Quantized Bits} & {\textbf{Raw Comp. Rate} $\gamma_r$} & {\textbf{Adj. Comp. Rate} $\gamma_a$} \\
\midrule
gzip (baseline)  & N/A & - & 0.336 & 0.336 \\
\midrule
\multirow{9}{*}{Llama-3.2-1B}
& None  & -  & 0.085 & 2.557 \\[0.5ex]
\cline{2-5}
& \multirow{4}{*}{GPTQ} & 8 & 0.085 & 1.321 \\
&                       & 4 & 0.087 & 0.705 \\
&                       & 3 & 0.096 & 0.560 \\
&                       & 2 & 0.185 & 0.494 \\[0.5ex]
\cline{2-5}
& \multirow{4}{*}{HQQ}  & 8 & 0.085 & 1.321 \\
&                       & 4 & 0.087 & 0.704 \\
&                       & 3 & 0.093 & 0.556 \\
&                       & 2 & 0.216 & 0.525 \\
\midrule
\multirow{9}{*}{Qwen2-0.5B}
& None  & -  & 0.104 & 1.092 \\[0.5ex]
\cline{2-5}
& \multirow{4}{*}{GPTQ} & 8 & 0.105 & 0.599 \\
&                       & 4 & 0.106 & 0.353 \\
&                       & 3 & 0.114 & 0.299 \\
&                       & 2 & 0.232 & 0.355 \\[0.5ex]
\cline{2-5}
& \multirow{4}{*}{HQQ}  & 8 & 0.104 & 0.598 \\
&                       & 4 & 0.106 & 0.353 \\
&                       & 3 & 0.109 & 0.294 \\
&                       & 2 & 0.138 & 0.262 \\
\midrule
\multirow{9}{*}{SmolLM-360M}
& None  & -  & 0.100 & 0.824 \\[0.5ex]
\cline{2-5}
& \multirow{4}{*}{GPTQ} & 8 & 0.100 & 0.462 \\
&                       & 4 & 0.103 & 0.284 \\
&                       & 3 & 0.115 & 0.251 \\
&                       & 2 & 0.403 & 0.493 \\[0.5ex]
\cline{2-5}
& \multirow{4}{*}{HQQ}  & 8 & 0.100 & 0.462 \\
&                       & 4 & 0.101 & 0.282 \\
&                       & 3 & 0.107 & 0.242 \\
&                       & 2 & 0.174 & 0.264 \\
\midrule
\multirow{9}{*}{SmolLM-135M}
& None  & -  & 0.114 & 0.383 \\[0.5ex]
\cline{2-5}
& \multirow{4}{*}{GPTQ} & 8 & 0.114 & 0.249 \\
&                       & 4 & 0.118 & 0.186 \\
&                       & 3 & 0.136 & 0.186 \\
&                       & 2 & 0.306 & 0.340 \\[0.5ex]
\cline{2-5}
& \multirow{4}{*}{HQQ}  & 8 & 0.114 & 0.249 \\
&                       & 4 & 0.117 & 0.184 \\
&                       & 3 & 0.129 & 0.180 \\
&                       & 2 & 0.257 & 0.291 \\
\bottomrule
\end{tabular}
\end{adjustbox}
\vspace{0.25cm}
\caption{\textbf{Compression Rates Under Different Quantization Schemes.} Shown above are the raw and adjusted compression rates for each of the models on the enwik9 dataset ($1$GB). The adjusted compression rate $\gamma_a$ accounts for the model size (see Section \ref{exp:setup:metrics}). Notice how quantization at $3$-bits still retains a good balance between compression rate and model size. At $2$-bit quantization, the model's predictive performance deteriorates significantly.}
\label{table:quantization}
\end{table}

\subsection{Compression Performance on Different Data Types}

\subsubsection{Non-English Data}
In this section, we evaluate the performance of LLM compression on various language sources. We include Gzip as a baseline comparison, and we evaluate the raw compression rate and the adjusted compression rate using $3$-bit HQQ quantization.

Of the languages listed above, Llama-3.2-1B was reported to be optimized for English, French, Spanish, and Thai. Therefore, we expect the compression rates for these languages to be exceed those of the other languages. Qwen-2-0.5B is an LLM trained on primarily English and Simplified Chinese data. Hence, their performance on English, Traditional Chinese, and possibly Japanese (since there are overlapping characters between Japanese and Traditional Chinese) should be better. SmolLM models were trained on primarily synthetic English data. It is expected that their performance on non-English data to drop by a large margin.

\begin{figure}
    \centering
    \includegraphics[width=\linewidth]{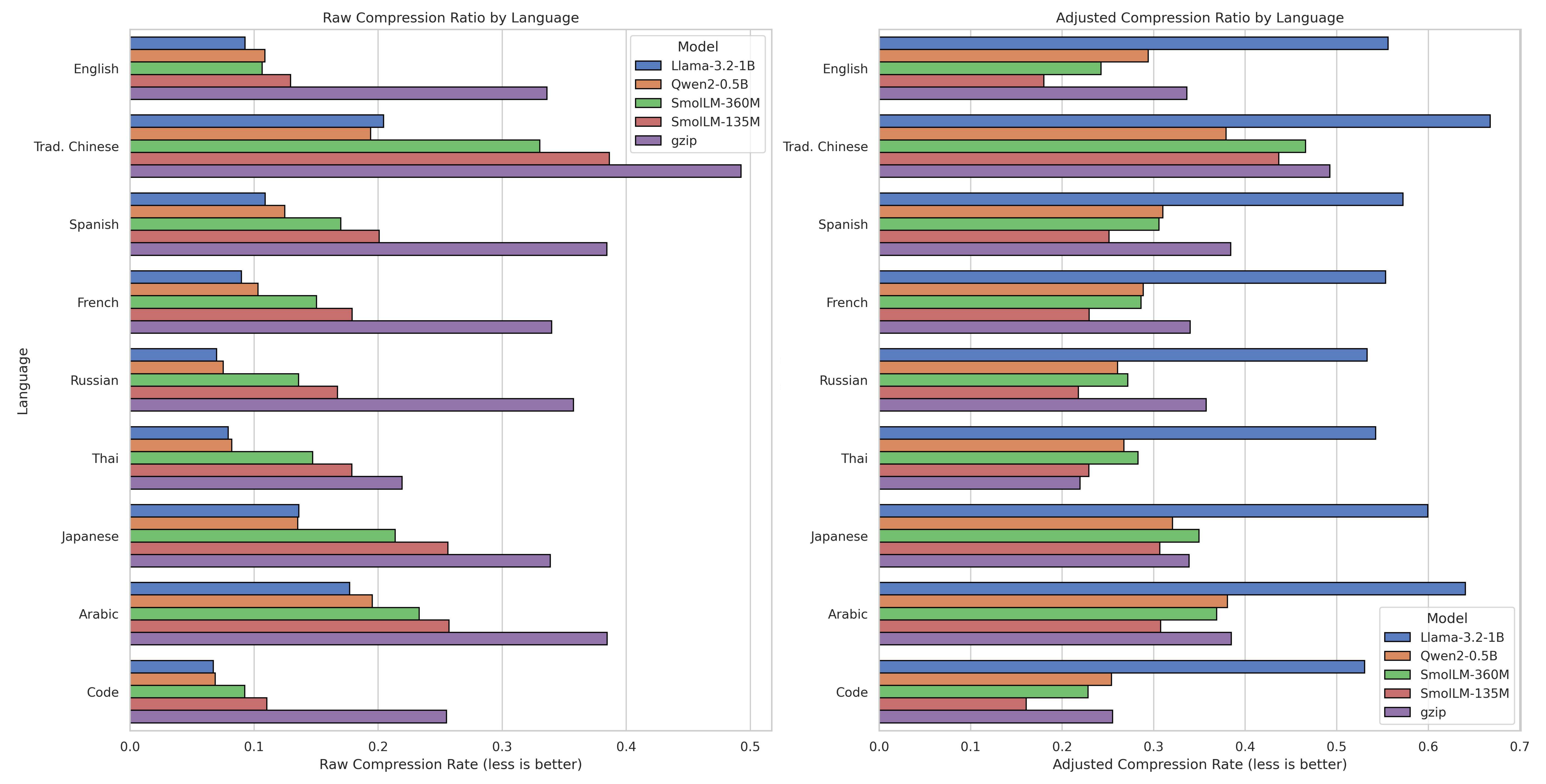}
    \caption{\textbf{Compression Rate on Different Languages.} Shown above is the compression rates of LLM compression on different languages. The performance of LLM compress is largely determined by amount of exposure the LLM has during pre-training. We incldue Gzip performance as baselinse. The raw compression rate is presented on the left, and the adjusted compression rate is given on the right.}
    \label{fig:exp:language_comp}
\end{figure}

The results of our experiment are presented in Figure \ref{fig:exp:language_comp}. The first observation is that Llama-3.2-1B able to achieve $\gamma_r < 0.12$ for all the languages it was trained on. In addition, Llama-3.2-1B was able to compress Russian text on a competitively level even though it was not trained on it. In contrast, Traditional Chinese and Arabic texts seems to requires more effort compressing.

Qwen-2-0.5B models seem to perform on a similar level as Llama-3.1-1B. We can also notice that Qwen-2-0.5B was able surpass Llama-3.2-1B in compressing Traditional Chinese and Japanese texts. This aligns with our hypothesis stated earlier.

SmolLM models yield a strong compression rate compared to Llama-3.2-1B and Qwen-2-0.5B on English texts. However, their performance on texts that do not share the same character set as English seemingly drops by a significant margin (i.e., Traditional Chinese, Russian, Thai, Japanese). This performance drop can be largely alluded to the limited training data during model pre-training. 

Of all language types, LLM's seem to handle code compression with ease. This is most likely due to the amount of code data that was used to pre-train these LLM's. 

\subsection{Byte Stream Data}
In this section, we evaluate the different methods to compress non-text byte stream data. Byte stream data are  essentially encoded data representations for particular file formats. Here, we would like to explore the question whether encoded byte stream data can be compressed just as well as text data. In Sections \ref{met:tok:byte_as_text}, \ref{met:tok:byte_as_int}, and \ref{met:tok:byte_as_byte}, we presented three different methods to tokenize byte stream data before passing them to the LLM for prediction. We follow the proposed methods to evaluate the compression performance of LLM's on PDF data encoded in their byte format.

\begin{figure}
    \centering
    \includegraphics[width=0.87\linewidth]{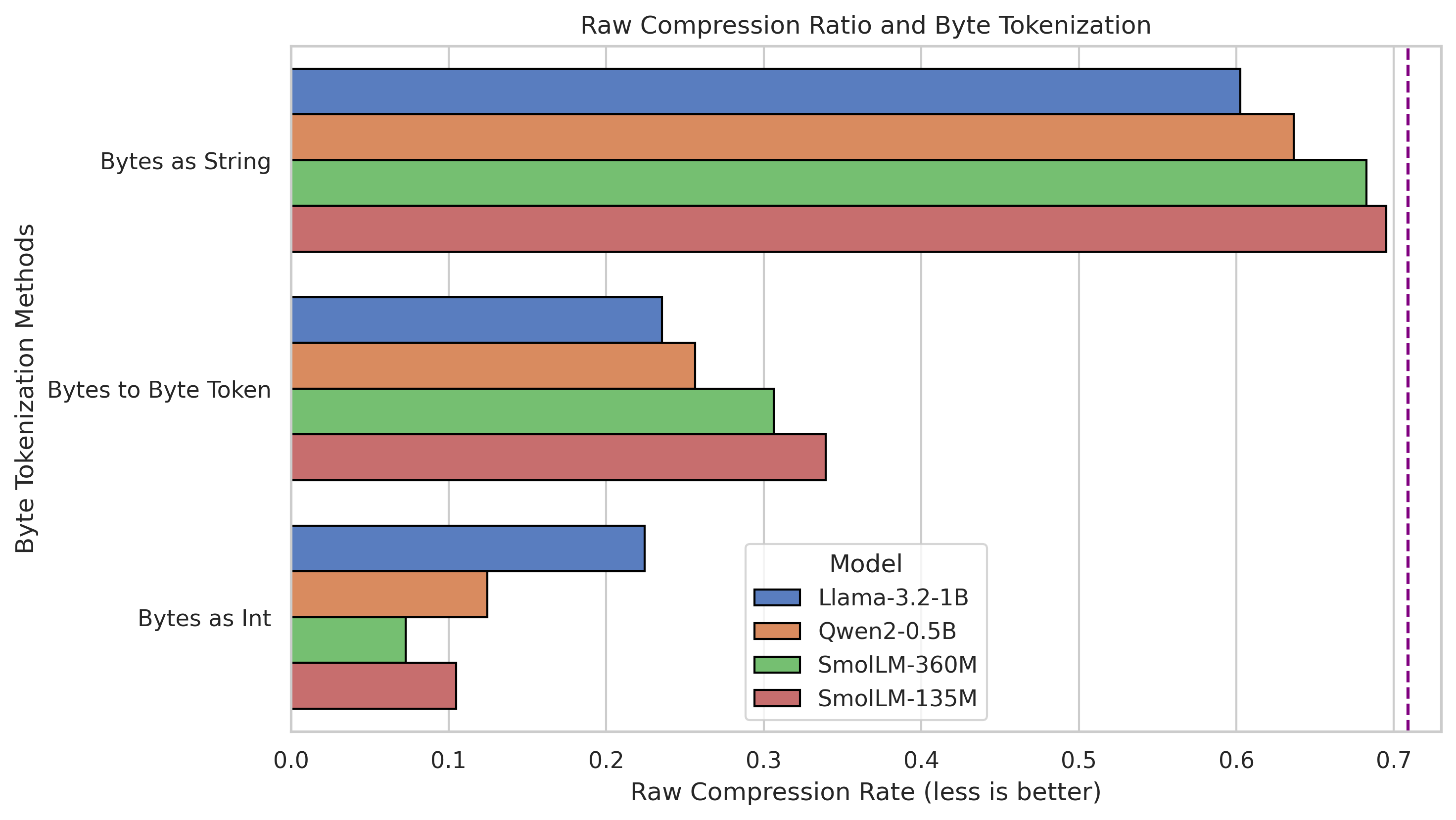}
    \caption{\textbf{Raw Compression Rate Under Different Byte Tokenizations.} Shown above are the raw compression rates three different byte tokenization methods. Notice that reading the byte encodings as raw texts (Section \ref{met:tok:byte_as_text}) yield performance very similar to standard compression (Gzip). Tokenizing bytes using their mapped byte tokens (see Section \ref{met:tok:byte_as_byte}) yields significant improvements. Maximum compression can be achieved by reading the bytes as integers (Section \ref{met:tok:byte_as_int}).}
    \label{fig:exp:byte_raw}
\end{figure}

\begin{figure}
    \centering
    \includegraphics[width=0.87\linewidth]{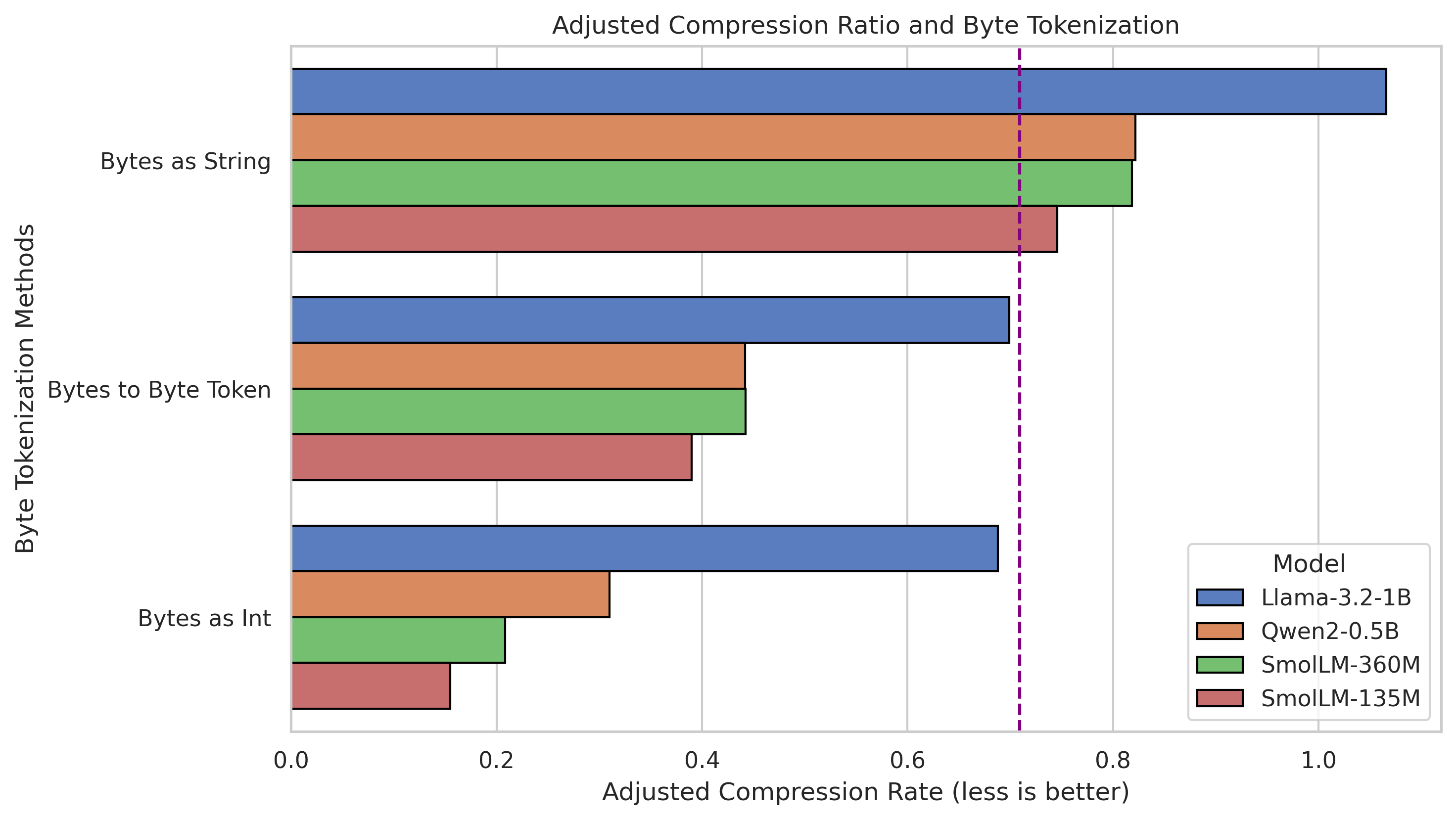}
    \caption{\textbf{Adjusted Compression Rate Under Different Byte Tokenizations.} Shown above are the adjusted compression rates three different byte tokenization methods. When accounting for the additional model weight, treating the byte stream as text (Section \ref{met:tok:byte_as_text}) offers no advantage compared to standard compression methods (i.e., Gzip). In contrast, processing the bytes as integers (Section \ref{met:tok:byte_as_int}) or their mapped byte tokens (Section \ref{met:tok:byte_as_byte}) results in significant improvements.}
    \label{fig:exp:byte_adj}
\end{figure}

The raw compression rate is presented in Figure \ref{fig:quant:raw_comp_ratio}. We can see that treating the byte stream as texts (Section \ref{met:tok:byte_as_text}) offers only a slight advantage over a baseline compressor (i.e. Gzip). In contrast, processing the byte stream as their mapped byte tokens (Sec \ref{met:tok:byte_as_byte}) offers a more competitive performance. The best compression rate can be achieved by mapping each byte to their integer representation in the tokenizer (Section \ref{met:tok:byte_as_int}). 

When considering the model size as part of the compression, we can see that treating the byte stream as texts (Section \ref{met:tok:byte_as_text}) is worse compared to the baseline compressors. To achieve the best compression, one should handle the byte stream as integers (Section \ref{met:tok:byte_as_int}) to achieve strong compression performance.

\section{Conclusions}
\label{sec:conclusions}

In this work, we revisit the topic of LLM compression from different perspectives. First, we considered methods to improve the adjusted compression rate by means of context length extension and model size reduction. Our experiments demonstrate that while longer context lengths offer an improvement in compression performance, the incremental improvement does not yield significant gains in reducing the adjusted compression rate. We considered two different LLM compression methods, GPTQ and HQQ, to reduce the model weights. Our experiments demonstrate that by reducing the model weight representations from their original $16$-bit representation down to a $3$-bit representation results in only a slight drop in raw compression rate. The benifit of such compression allows for the model size to be compressed significantly. This configuration allows us to compress a un-trained model that achieves the SOTA adjusted compression rate of $18\%$ on the enwik9 \cite{hutter2006prize} dataset. 

Furthermore, we explored the use of compressing non-English data that spans over a range of languages as well as code data. We show that LLM's trained on specific languages are often correlated with better compression performance. In addition to text data, we experimented with compressing byte stream data from PDF encoded files. We presented several possible methods to process the PDF encoded byte streams as tokens. Our experiments show that processing the byte streams as integer representations in the tokenizer results in the best compression. 

Some possible future works may include compressing additional data formats. Potential for compressing deep neural network model weights may also be explored. We hope this work highlights the importance of LLM in data compression, and we believe that future data compressors may involve more efficient LLM's specialized for compression.

\bibliographystyle{alpha}
\bibliography{references}

\end{document}